\documentclass[conference]{IEEEtran}
\usepackage{cite}
\usepackage{amsmath,amssymb,amsfonts}
\usepackage{algorithmic}
\usepackage{graphicx}
\usepackage{textcomp}
\usepackage{xcolor}
\usepackage{listings}
\usepackage{verbatim}
\usepackage{tcolorbox}
\usepackage{tikz}
\usepackage{multirow}
\usepackage{pgfplots}
\usepackage{url}
\pgfplotsset{compat=1.18}
\usepackage{footnote}
\usepackage{threeparttable}
\usetikzlibrary{shapes.geometric, arrows, fit, positioning, calc, arrows.meta}

\def\BibTeX{{\rm B\kern-.05em{\sc i\kern-.025em b}\kern-.08em
    T\kern-.1667em\lower.7ex\hbox{E}\kern-.125emX}}

\begin{document}

\title{ColBERT Retrieval and Ensemble Response Scoring for Language Model Question Answering\\
}

\author{\IEEEauthorblockN{Alex Gichamba\IEEEauthorrefmark{1}, Tewodros Kederalah Idris\IEEEauthorrefmark{1}, Brian Ebiyau\IEEEauthorrefmark{1}, Eric Nyberg\IEEEauthorrefmark{2}, Teruko Mitamura\IEEEauthorrefmark{2}}
\IEEEauthorblockA{\IEEEauthorrefmark{1} Carnegie Mellon University Africa, Kigali, Rwanda}
\IEEEauthorblockA{\IEEEauthorrefmark{2}Carnegie Mellon University, Pittsburgh, USA\\
Email: \{angicham, tidris, bebiyau, en09, teruko\}@andrew.cmu.edu}
}

\maketitle

\begin{abstract}
Domain-specific question answering remains challenging for language models, given the deep technical knowledge required to answer questions correctly. This difficulty is amplified for smaller language models that cannot encode as much information in their parameters as larger models. The ``Specializing Large Language Models for Telecom Networks" challenge aimed to enhance the performance of two small language models, Phi-2 and Falcon-7B in telecommunication question answering. In this paper, we present our question answering systems for this challenge. Our solutions achieved leading marks of 81.9\% accuracy for Phi-2 and 57.3\% for Falcon-7B.  We have publicly released our code and fine-tuned models\footnote{\path{https://github.com/Alexgichamba/itu_qna_challenge}}.

\end{abstract}


\section{Introduction}
Advances in Large Language Models (LLMs) have significantly enhanced their performance across various Natural Language Processing (NLP) tasks. LLM series such as OpenAI's GPTs \cite{brown_language_2020,openai_gpt-4_2024}, have grown substantially in both model parameters and training data volume, driving much of this progress. One notable area of improvement is open-domain question answering (QA) \cite{hendrycks_measuring_2021}. Pre-trained on a wide range of domain-general data, LLMs excel in leveraging common knowledge \cite{yang_empower_2023}. However, this strength does not fully extend to domain-specific QA, which demands specialized knowledge  often underrepresented in training datasets \cite{yang_empower_2023,drazen_benefits_2023}. Furthermore, LLMs can sometimes produce responses that conflict with established facts \cite{huang_survey_2023}, posing a significant challenge in domain-specific QA where factual accuracy is crucial.

The ``Specializing Large Language Models for Telecom Networks" challenge invited solutions for specializing two small language models, Phi-2 \cite{noauthor_phi-2_nodate} and Falcon-7B \cite{almazrouei2023falcon} to answer multiple-choice questions on telecommunication standards. The challenge featured two tracks, one for each language model, with accuracy serving as the performance metric. Participants were provided with 1,461 standards questions from TeleQnA \cite{maatouk_teleqna_2023} for training purposes, along with 3rd Generation Partnership Project (3GPP) technical documents for information retrieval. The evaluation dataset consisted of 2,366 questions, 366 of which were sourced from TeleQnA and served as the public evaluation set. The remaining 2,000 questions had not been publicly released at the time of the competition. These questions covered a broad range of subdomains within telecommunications, including propagation, media standards, and networking. Technical terms and abbreviations from these subdomains posed challenges for non-specialized language models. Some questions required reasoning across all options, especially when the initial options were either all true or all false.

\begin{figure}[t]
    \centering
    \includegraphics[width=\linewidth]{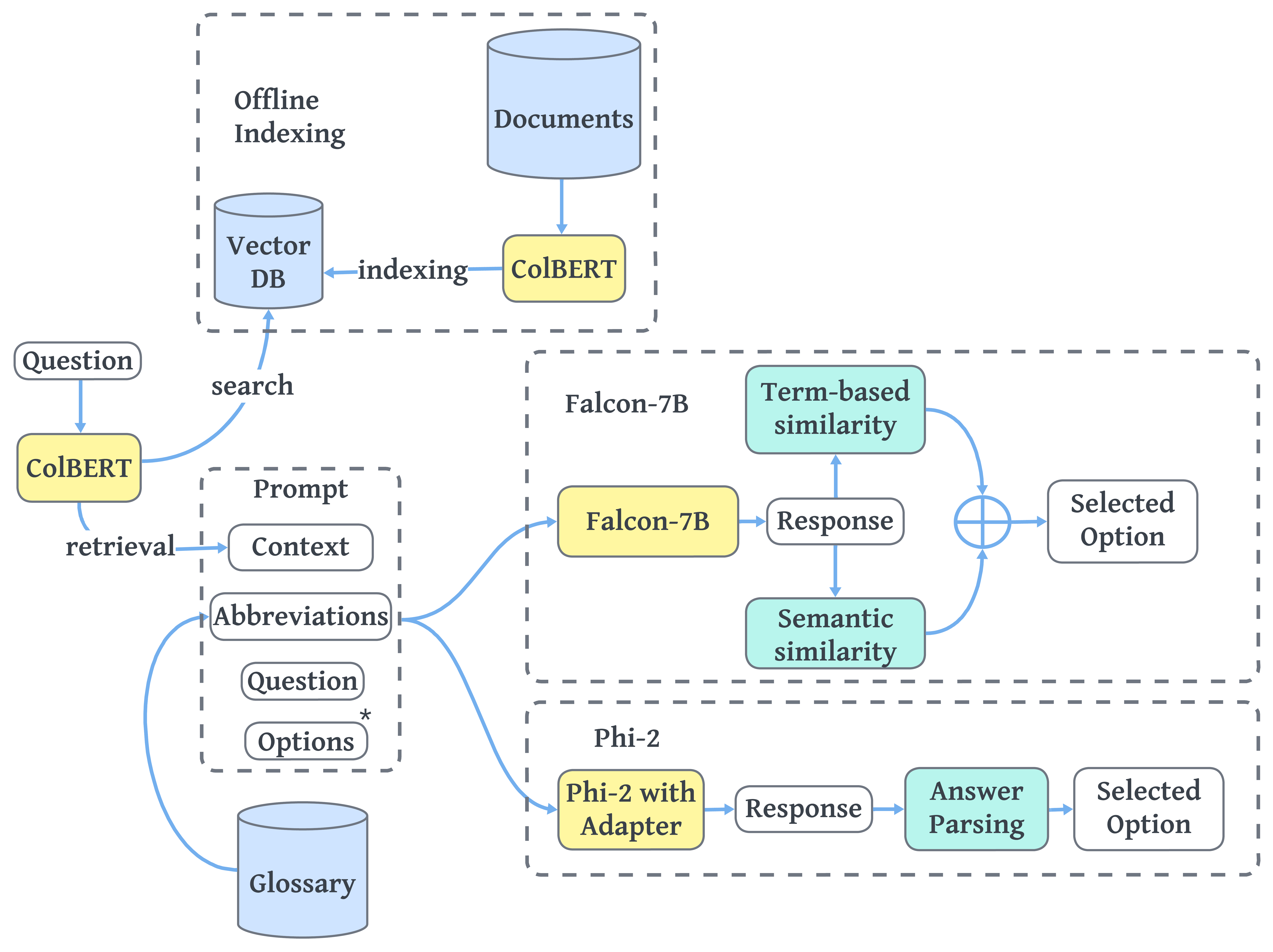}
    \caption{Both our QA systems feature a ColBERT-based retrieval pipeline and lexicon enhancement. Phi-2 is fine-tuned for instruction alignment and to invoke better reasoning.\\ \textbf{*} The Falcon-7B prompt doesn't include the options. Since the responses are not conditioned on the options, we evaluated them using an ensemble scoring system to find the most likely option.}
    \label{fig:combined_sys}
\end{figure}

Given the nature of the questions and the constraints imposed by this challenge, several significant hurdles emerged. Both Phi-2 and Falcon-7B are relatively small compared to current state-of-the-art models, which limits their ability to encode the necessary domain-specific knowledge within their weights. We implemented baseline solutions for each track, providing only brief instructions along with the questions and options in the prompt. The baseline solutions achieved accuracies of 41. 5\% for Phi-2 and 24.5\% for Falcon-7B. From these baseline systems, we observed that the language models frequently failed to adhere to the instructions provided in the prompts. In particular, Phi-2 often generated responses that did not conform to the instructed output format. Similarly, Falcon-7B exhibited notable difficulty in reasoning through questions when options were included in the prompt. This limitation has also been reported in previous works \cite{tas-24,almazrouei2023falcon}.

In this paper, we describe our QA systems based on Phi-2 and Falcon-7B. The system diagrams are shown in \figurename~\ref{fig:combined_sys}. We developed a Retrieval-Augmented Generation (RAG) pipeline based on ColBERT \cite{santhanam_colbertv2_2022} to provide relevant context from the documents for each query at inference time. In addition, we built a glossary of technical abbreviations from the 3GPP documents and included the relevant abbreviations and their full forms in the prompt for each question. For Phi-2, we fine-tuned a LoRA adapter \cite{hu_lora_2021} on the training questions with a prompt that included the retrieved context. For Falcon-7B, we let the language model generate responses based only on the question, context, and abbreviations, and then developed a scoring mechanism to determine the most likely option based on the generated response.

Our solutions proved highly effective, achieving the highest accuracies in their respective tracks: 81.9\% for Phi-2 and 57.3\% for Falcon-7B on the private evaluation dataset. The performance of our Falcon-7B QA system is a significant improvement over previously reported results in the domain \cite{tas-24} and in multiple-choice QA in general \cite{almazrouei2023falcon}. 

\section{Related Work}


The application of LLMs in highly specialized domains, such as telecommunications, presents unique challenges due to complex and dynamic documentation \cite{bornea_telco-rag_2024}. Ahmed et al. \cite{tas-24} evaluated the performance of LLMs in the telecommunications domain, focusing on zero-shot evaluations of models like Llama-2, Falcon, Mistral, and Zephyr. These models, more resource-efficient than ChatGPT, demonstrated comparable performance to fine-tuned models, underscoring the benefits of extensive pre-training. TelcoRAG \cite{bornea_telco-rag_2024}, an open-source RAG framework, was developed to address the challenges in processing telecommunications standards, specifically from the 3GPP. This framework provides a foundation for applying RAG in telecommunications and other technical fields, offering valuable guidelines for deploying RAG pipelines on specialized content. The study \cite{bornea_telco-rag_2024} also provides a foundation for using LLMs for QA tasks in technical domain, particularly telecommunications.

RAG has emerged as a significant approach to enhancing LLM performance in knowledge-intensive tasks. Recent advancements in dense retrieval methods, such as ColBERT\cite{santhanam_colbertv2_2022} and the FAISS\cite{Johnson_2021} library, have been instrumental in improving RAG systems by enabling efficient similarity search for large-scale documents. Textual entailment \cite{dagan_pascal_2006}, a crucial aspect of QA systems, has evolved from rule-based methods to machine learning and neural-based approaches. The NTCIR-9 RITE task \cite{shima_overview_2011} provided a framework to evaluate textual entailment in multiple languages, laying the foundation for future work in the field.
Hendrycks et al. \cite{hendrycks_measuring_2021} introduced a benchmark to assess LLM multitask accuracy across 57 diverse tasks, demonstrating that even the most advanced models, such as GPT-3, only surpass random chance by 20 percentage points on average. This performance gap highlights the need for domain-specific datasets. Bornea et al. \cite{bornea_telco-rag_2024} utilized the Telco-RAG framework to enhance LLM performance on QA tasks in the telecommunication domain, particularly for the TeleQnA dataset. The study demonstrated that domain-specific adaptations can significantly improve the performance of LLMs in specialized QA tasks.
Neural-based models now outperform rule-based methods by capturing deeper semantic nuances \cite{alharahseheh_survey_2022}. This study employs ColBERT-based RAG appraoch for retrieving dense representations. The study also uses neural-based models alongside statistical methods like TF-IDF and word overlap for text entailment.


\section{Methodology}

Our approach is tailored to address the lack of knowledge in telecommunications, nonadherence to instructions, and lexicon mismatch. We developed distinct methodologies, but with some shared components, for the Phi-2 and Falcon-7B tracks.

\subsection{Phi-2}

For the Phi-2 track, we implemented a comprehensive system that integrates fine-tuning, RAG, and domain-specific lexicon enhancement. \figurename~\ref{fig:combined_sys} illustrates the architecture of our Phi-2 QA system, demonstrating the interplay between these components.

\subsubsection{Fine-tuning}

We used LoRA to fine-tune Phi-2 on the training questions. We had two primary objectives. First, we sought to align the language model with the output format and prompt expected at inference time. The prompts included instructions, questions, multiple-choice options, relevant abbreviations, and context. Second, we aimed to improve the model's reasoning capabilities. Instead of training the model to predict the correct answers alone, we expanded the targets to include explanations. We expected that including explanations as prediction targets could provide insight into the model's decision-making process.

\subsubsection{Retrieval-Augmented Generation (RAG)}

Our RAG pipeline leverages 554 3GPP standard technical documents. From empirical results in \figurename~\ref{fig:accuracy_vs_colbert_k}, we determined that segmenting these documents into chunks of 150 tokens proved crucial in maintaining the semantic coherence of technical concepts while allowing precise information retrieval.

For the retrieval component, we used the ColBERT model \cite{santhanam_colbertv2_2022}. ColBERT's ability to capture intricate meaning between terms is particularly beneficial in the highly technical telecommunication domain. The resulting vectorized chunks form the foundation of our retrieval system.

During the QA process, our system utilizes FAISS [18] for efficient similarity search. From our experiments with the number of retrieved chunks ($k$) recorded in \figurename~\ref{fig:accuracy_vs_colbert_k}, we observed a general trend of improved performance as $k$ increased, although this improvement was not strictly monotonic. However, our choice of $k$ was ultimately limited by Phi-2's context-window limitation of 2048 tokens. This restriction led us to settle on a value of $k=13$ as the optimal choice within these constraints.

\subsubsection{Technical Abbreviation Expansion}

To address the challenge of domain-specific terms that may be previously unseen by the model, we developed an abbreviation expansion mechanism. We systematically extracted and compiled a glossary of technical abbreviations from the 3GPP standard documents.

For each question, we implement a dynamic abbreviation expansion process. This involves analyzing the question text and options to identify technical abbreviations, querying our compiled glossary to retrieve the corresponding full forms, and incorporating these expansions into the prompt presented to the language model.

\subsubsection{Prompt Engineering}

The structured prompt includes the following.

\begin{itemize}
    \item Unambiguous instructions for the task
    \item Retrieved context from relevant documents
    \item Expanded abbreviations and definitions from our glossary
    \item The question and multiple-choice options
\end{itemize}

This comprehensive prompt structure enables Phi-2 to leverage both its fine-tuned knowledge and dynamically retrieved information to generate accurate responses.

\subsection{Falcon-7B}

For the Falcon-7B track, we adopt the same retrieval pipeline as used for Phi-2, the only difference being the number of chunks used as context. Unlike Phi-2 where we used as many chunks as the context window allowed, for Falcon-7B we found that providing more than 3 chunks in the prompt degrades the performance as shown in ~Table~\ref{tab:falcon_k}. We thus use 3 chunks in the final system. Following our Phi-2 approach, we also address the lexical mismatch by expanding the abbreviations found in the query and options. The key differences between the two systems are the prompts used and the scoring mechanism developed for the responses generated by Falcon-7B. The final system architecture is shown in \figurename~\ref{fig:combined_sys}.

\subsubsection{Prompting without Options}
From our initial experiments on Falcon-7B we found that the model struggles to reason when presented with options for multiple choice QA. We found no apparent ability of the language model to use the question, context, and options to reason toward an informed response. Therefore, we do not provide the options in the prompt. With this change alone, we found that the LLM can use the context to generate a relevant answer to the query.

\subsubsection{Response Scoring}
\label{sub:scoring_response}
Without options in the prompt, we found that, in the best case, the response is a verbose version of the correct option, often reflecting the most relevant part of the context. This is observed when the answer is found within the top 3 retrieved chunks and the question is unambiguous. Going forward, we refer to this as the best-case response regarding our Falcon-7B QA system. We thus developed a scoring method that finds the most likely option given the response. We hypothesize that this task most closely resembles that of textual entailment or natural language inference \cite{alharahseheh_survey_2022}. We found that the wrong options are often relevant but contradict the correct option, and so we formulate this as a task of textual entailment recognition without the neutral case.

To quickly test this hypothesis, we chose to use the cosine similarity between the embeddings of the response and the options as a measure of semantic similarity. We developed a fine-tuning objective for the embedding model that seeks to maximize the similarity between texts that entail one another and minimize the similarity between contradicting texts. We select the large General Text Embedding (GTE) v1.5 model \cite{li_towards_2023} and fine-tune it with triplet loss. Given the limited training set permitted for the challenge, we found that the best proxy for the responses seen at inference time to be the explanations for the correct options in the training set. For each question, we selected the explanation as the anchor, the correct option as the positive instance, and a randomly selected option as the negative instance.



After fine-tuning, we noticed the improved ability of the embedding model to match the relevant responses to the correct option. However, we hypothesized that this could be further improved by incorporating some measure of term overlap. We compute a Term Frequency-Inverse Document Frequency (TF-IDF) weighted word overlap score between a given response and the set of options. We sum the product of TF-IDF scores for shared words between each option and the response. This approach gives more weight to significant shared terms while reducing the impact of common, less informative words.

The final score is a weighted sum of the term overlap and the cosine similarity. We run ablations for different static weights and select $ \alpha_1 = 0.2 $ and $ \alpha_2 = 0.8 $, where:

\begin{itemize}
    \item $\alpha_1$ is the static weight for the term overlap score
    \item $\alpha_2$ is the static weight for the cosine similarity
\end{itemize}




\section{Experiments and Results}
In this section, we present the experiments conducted to evaluate the performance of our design choices. All experiments were performed on the public subset of the dataset.
\subsection{Phi-2 Track: Retrieval Method Exploration and Parameter Sensitivity}

We conducted a series of experiments to identify the most suitable retrieval method and optimize its parameters. We evaluated the following three main retrieval approaches: dense retrieval using ColBERT \cite{santhanam_colbertv2_2022}, a state-of-the-art dense retrieval model; sparse retrieval using BM25 \cite{robertson1995okapi}, a traditional sparse retrieval method, and an ensemble approach combining both ColBERT and BM25.

We hypothesized that the ensemble approach could take advantage of the strengths of both sparse and dense retrieval, potentially yielding superior results in the highly technical telecommunication domain, particularly when working with the smaller Phi-2 model.

\subsubsection{Parameter Sensitivity in Phi-2 Setup}

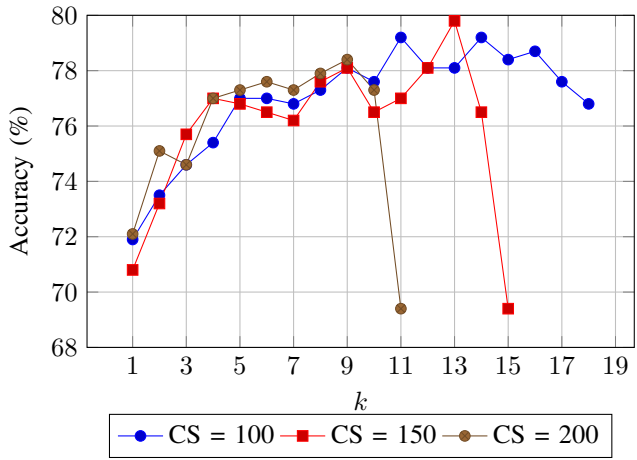
\begin{figure}[t]
    \centering
    \begin{tikzpicture}
        \begin{axis}[
            xlabel={$k$},
            ylabel={Accuracy (\%)},
            xtick=data,
            grid=major,
            ymin=68, ymax=80,
            xtick={1,3,5,7,9,11,13,15,17,19},
            width=\linewidth,
            height=6cm,
            ytick={68, 70, 72, 74, 76, 78, 80},
            legend style={at={(0.5,-0.2)}, anchor=north, legend columns=-1}
        ]

    \addplot coordinates {
        (1,71.9)
        (2,73.5)
        (3,74.6)
        (4,75.4)
        (5,77.0)
        (6,77.0)
        (7,76.8)
        (8,77.3)
        (9,78.1)
        (10,77.6)
        (11,79.2)
        (12,78.1)
        (13,78.1)
        (14,79.2)
        (15,78.4)
        (16,78.7)
        (17,77.6)
        (18,76.8)
    };
    \addlegendentry{CS = 100}

        \addplot coordinates {
            (1,70.8)
            (2,73.2)
            (3,75.7)
            (4,77.0)
            (5,76.8)
            (6,76.5)
            (7,76.2)
            (8,77.6)
            (9,78.1)
            (10,76.5)
            (11,77.0)
            (12,78.1)
            (13,79.8)
            (14,76.5)
            (15,69.4)
        };
        \addlegendentry{CS = 150}

        \addplot coordinates {
            (1,72.1)
            (2,75.1)
            (3,74.6)
            (4,77.0)
            (5,77.3)
            (6,77.6)
            (7,77.3)
            (8,77.9)
            (9,78.4)
            (10,77.3)
            (11,69.4)
        };
        \addlegendentry{CS = 200}

        \end{axis}
    \end{tikzpicture}
\caption{Accuracy as a function of the number of chunks ($k$) for different chunk sizes (CS). The number of chunks was incrementally increased beyond the context window length, leading to sharp declines for larger values of $k$.}

    \label{fig:accuracy_vs_colbert_k}
\end{figure}

Throughout our experimentation with Phi-2, we observed a high degree of sensitivity to various parameters. \figurename~\ref{fig:accuracy_vs_colbert_k} summarizes our findings for different chunk sizes and top-k values using ColBERT retrieval with Phi-2.


Key observations from our parameter sensitivity analysis for Phi-2 include:

\begin{itemize}    
    \item \textbf{Chunk Size:} The size of document chunks used for indexing played a crucial role in Phi-2's performance. We found that a chunk size of 150 tokens provided the best balance between context preservation and retrieval precision for this model.
    
    \item \textbf{Input Token Limitations:} The constrained input token limit of Phi-2 (2048 tokens) directly impacted our ability to scale the number of retrieved passages and chunk sizes. As observed in \figurename~\ref{fig:accuracy_vs_colbert_k} the performance drops steeply when our inputs exceed the context window's length.
\end{itemize}

\subsubsection{Effect of fine-tuning with context} We also investigated the effect of varying the number of top-k chunks used as context during fine-tuning. We fine-tuned three LoRA adapters with different $ k $ values: $ k = {0, 3, 7} $, each corresponding to a separate training run. For inference, we maintained a constant $ k = 13 $. As we increased $k$ we observed a $0.8\%$ to $0.9\%$ improvement in accuracy. We found the top-k value used for inference to be much more critical in driving the performance as seen in \figurename~\ref{fig:accuracy_vs_colbert_k}.







\subsubsection{BM25 and ColBERT Ensemble}

For the ensemble approach with Phi-2, we experimented with combining results from both BM25 and ColBERT retrievers. Our method involved selecting top-k chunks from each retriever to construct the context for the language model.

Contrary to our initial hypothesis, the ensemble approach did not outperform the use of ColBERT alone with Phi-2. To understand why, we investigated the relevance of the retrieved chunks for both methods. Given the labor-intensive nature of human evaluation, we initially employed an LLM to assess recall at 13 (the optimal number of chunks for Phi-2), prompting it to determine whether each passage was sufficient to answer a question. Using GPT-4o-mini \cite{gpt-4o_nodate}, we found that for 96.99\% of the questions in the public evaluation set, the correct answer appeared within the top 13 retrieved chunks. However, upon closer inspection of the results, we found this number to be grossly overestimated due to a significant number of false positives.

To obtain a more accurate estimate, we randomly sampled 122 questions from the public evaluation set and manually evaluated the binary recall for the ColBERT and BM25 retrieval systems. In ~Table~\ref{tab:bin_recall} we report the results of this evaluation. 

\begin{table}[ht]
\centering
\caption{Proportion of Questions where an Answer-Bearing Chunk was Found in the Top 13 Retrieved Chunks}
\label{tab:bin_recall}
\begin{tabular}{|c|c|c|}
\hline
Method & Number of Questions & Percentage of Questions \\
\hline
ColBERT & 98 & 80.3\% \\
BM25 & 94 & 77.1\% \\
\hline
\end{tabular}
\end{table}

This finding provides valuable insights into the nature of our dataset and the effectiveness of dense retrieval methods in our specific context:

\begin{itemize}
    \item \textbf{Lexical Similarity:} The superior performance of ColBERT alone suggests that there was no  significant lexical gap between the queries (questions) and the relevant documents in our dataset. A lexical gap would be expected if the queries used substantially different vocabulary or phrasing compared to the document content. The absence of this gap is likely due to our evaluation set being drawn from the same distribution as the training data, ensuring consistent terminology and phrasing.

    \item \textbf{Semantic Understanding:} ColBERT's effectiveness in our domain-specific task underscores its ability to model semantic relationships; it can recognize subtle distinctions between words and phrases, even when working with a smaller model like Phi-2. This suggests that semantic understanding was more crucial than handling lexical variations in our dataset.

    \item \textbf{Ensemble Limitations:} The ensemble approach, which we initially hypothesized would help bridge potential lexical gaps, did not provide additional benefits. This further corroborates the absence of significant lexical variations between queries and relevant documents in our evaluation set.

    \item \textbf{Context Quality vs. Quantity:} As we increased the number of chunks retrieved from BM25 in our ensemble, we observed a deterioration in overall system performance. This indicates that the quality of retrieved context, particularly the semantic relevance captured by ColBERT, was more important than the quantity of keyword-matched chunks from BM25.
\end{itemize}

These findings reinforce the effectiveness of dense retrieval methods, particularly ColBERT, in capturing the intricate semantic patterns present in our domain-specific corpus. However, it is important to note that while this approach was optimal for our current dataset, the dynamics might change in a real-world deployment scenario where end-users formulate their own queries. In such cases, a wider variety of lexical variations may be encountered, potentially making an ensemble approach more beneficial.

\subsubsection{Prompt Engineering for Phi-2}

Our experiments with Phi-2 prompting strategies revealed that simpler prompts led to better model performance. We found that providing clear and concise instructions yielded superior results compared to more complex strategies such as few-shot prompting. 

\subsection{Falcon-7B Track: Context Utilization, Optimal Scoring Weights and Fine-tuning the Embedding Model}
\subsubsection{Context Utilization} We evaluated the ability of Falcon-7B to utilize the entire context window, as was demonstrated by Phi-2. We selected the optimal chunk size determined from the Phi-2 experiment (shown in \figurename~\ref{fig:accuracy_vs_colbert_k}) and increased the number of chunks added to the prompt. In ~Table~\ref{tab:falcon_k} it is observed that Falcon-7B is unable to disregard the noise introduced by irrelevant passages, or refer to relevant passages that are introduced later in the prompt.

\begin{table}[ht]
    \centering
    \caption{Accuracy (\%) for different top-k values of context provided in the Falcon-7B prompt}
    \begin{tabular}{|c|c|}
    \hline
    Number of chunks used as context, $ k$     & Accuracy (\%) \\
    \hline
    1       & 57.7 \\
    2       & 58.7 \\
    3       & \textbf{59.6} \\
    4       & 58.2 \\
    5       & 54.6 \\
    6       & 53.8 \\
    \hline
    \end{tabular}
    \label{tab:falcon_k}
\end{table}
\subsubsection{Optimal Scoring Weights} To determine the optimal weights for the term overlap and similarity scores, we uniformly varied the weights and scored the same set of responses. From the results in Table \ref{tab:scoring_weights}, we found that giving a higher weight to the similarity score is generally more beneficial. However, incorporating the overlap score with a small weight yields better accuracy than relying solely on the similarity score. Using the overlap score alone is the least effective, as it fails to capture the semantics needed to match the correct response to the appropriate option, which cannot be fully conveyed through shared terms alone.
\begin{table}[ht]
\centering
\caption{Accuracy (\%) for different scoring weights}
\label{tab:scoring_weights}
\begin{tabular}{|c|c|c|}
\hline
Overlap Weight, $\alpha_1$ & Similarity Weight, $\alpha_2$ & Accuracy \\
\hline
0.0 & 1.0 & 58.2\\
0.2 & 0.8 & \textbf{58.7}\\
0.4 & 0.6 & 56.3\\
0.6 & 0.4 & 53.3\\
0.8 & 0.2 & 52.5\\
1.0 & 0.0 & 46.7\\
\hline
\end{tabular}
\end{table}
\subsubsection{Enhancing the Embedding Model with an Entailment Objective}
We evaluate the performance of the large GTE \cite{li_towards_2023} embedding model before and after fine-tuning it with the entailment objective. We report the accuracy values from scoring the responses with only the cosine similarity scores from the embeddings. The accuracy is $47.3\%$ when scoring with the base model, and $58.2\%$ with the fine-tuned model.

To help validate that the fine-tuning objective would approximate learning textual entailment recognition, we apply the embeddings to an out-of-domain textual entailment recognition task. We select the development set of the MultiNLI \cite{williams-etal-2018-broad} dataset for this comparison. We mapped the labels to discrete ternary numerical target values, with the same range as cosine similarity then computed the mean absolute error (MAE) between the similarity scores and the targets. The MAE was $7\%$ lower for the fine-tuned model. These results show that fine-tuning with triplets where positive instances entail the anchors and negative instances contradict the anchors can be an objective to learn approximate soft scores for textual entailment recognition.

\subsection{Comparisons with leading LLMs}
We assessed the performance of our QA systems in comparison to several top LLMs. In these experiments, the models were not fine-tuned. Table~\ref{tab:diff_LLMs} (a) shows results for experiments where context, technical abbreviation expansions (TAEs), and options were included in the prompt. The highest accuracy was achieved by Llama 3.1 405B \cite{dubey_llama_2024}, outperforming our Phi-2 QA system by 6.3\%. In Table~\ref{tab:diff_LLMs} (b), we excluded context and TAEs while retaining the options to evaluate the models' inherent knowledge. Here, we observe that the larger LLMs significantly outperform the smaller LLM. We hypothesize that this is largely attributed to the larger training datasets used when scaling LLMs and the additional capacity of the models to encode relevant information in their parameters. Finally, Table~\ref{tab:diff_LLMs} (c) assesses the performance with context and TAEs included but without options. Llama 3.1 405B outperform the rest. The model appeared to be better than other LLMs at distilling their responses into short sentences that more closely resemble the correct options.
\begin{table}[ht]
\begin{threeparttable}
\centering
\caption{Accuracy (\%) for different LLMs on the public subset of the challenge evaluation dataset}
\label{tab:diff_LLMs}
\begin{tabular}{|c|c|c|c|c|}
\hline 
& LLM & Context + TAE \tnote{1} & Options & Accuracy \\
\hline
a) & Llama 3.1 405B \cite{dubey_llama_2024} & \checkmark & \checkmark & \textbf{86.6}\\
&Claude 3.5 Sonnet \cite{anthopic_introducing} & \checkmark & \checkmark & {84.7}\\
&GPT-4o-mini \cite{gpt-4o_nodate} & \checkmark & \checkmark & 83.3\\
&(Ours) Phi-2 & \checkmark & \checkmark & 80.3\\
\hline
b)&Claude 3.5 Sonnet \cite{anthopic_introducing}& $\times$ &\checkmark & \textbf{71.6}\\
&Llama 3.1 405B \cite{dubey_llama_2024} & $\times$ & \checkmark & 69.7\\
&GPT-4o-mini \cite{gpt-4o_nodate} & $\times$ & \checkmark & 65.3\\
&(Ours) Phi-2 & $\times$ & \checkmark & 55.2\\
\hline
c)&Llama 3.1 405B \cite{dubey_llama_2024} & \checkmark & $\times$ & \textbf{72.7}\\
&GPT-4o-mini \cite{gpt-4o_nodate} & \checkmark & $\times$ & {68.9}\\
&Claude 3.5 Sonnet \cite{anthopic_introducing} & \checkmark & $\times$ & 68.6\\
&(Ours) Falcon-7B & \checkmark & $\times$ & 60.1\\
\hline
\end{tabular}
     \begin{tablenotes}
       \item [1] TAE: Technical Abbreviation Expansion
     \end{tablenotes}
\end{threeparttable}
\end{table}

\subsection{Out of domain evaluation}
To assess the generalizability of our methods across different domains, we evaluated our QA systems on the pharmacology subset of the MedMCQA \cite{pal_medmcqa_2022} validation dataset. We selected pharmacology because it was one of the larger subsets for which we had access to a knowledge base of relevant textbooks. As presented in Table \ref{tab:ood}, the Phi-2 model fine-tuned on pharmacology questions achieved the highest accuracy. Notably, the Phi-2 model fine-tuned on telecommunication questions closely followed, with just a 1.7\% difference. These findings imply that our fine-tuning objectives primarily enhance instruction and task alignment, rather than knowledge alignment. The Falcon-7B QA system did not outperform the baseline, underscoring the importance of constraining reasoning through the provided options. Interestingly, the optimal scoring weights for this evaluation were the same as those for the telecommunication questions, despite most options in MedMCQA being shorter than those in the challenge dataset. This further highlights the robustness of the ensemble scoring system in multiple-choice QA.
\begin{table}[ht]
    \centering
    \begin{threeparttable}
        \caption{Accuracy (\%) on the pharmacology subset of the MedMCQA validation set}
        \label{tab:ood}
        \centering
        \begin{tabular}{|c|c|}
        \hline
        QA system  & Accuracy (\%) \\
        \hline
        (Ours) Phi-2 \tnote{1} & 56.4\\
        (Ours) Phi-2 \tnote{2} & 54.7\\
        (Baseline) PubMedBERT \cite{gu_domain-specific_2022} & 46.0\\
        (Ours) Falcon-7B & 38.7\\
        \hline
        \end{tabular}
        \begin{tablenotes}
        \centering
        \item [1] Fine-tuned on Pharmacology QA
        \item [2] Fine-tuned on Telecommunication QA
        \end{tablenotes}
    \end{threeparttable}
\end{table}

\section{Analysis}
\subsection{Inference Time}

To complement our accuracy results, we conducted timing analyses for the complete pipeline execution on an NVIDIA L4 GPU. Table~\ref{tab:qa-pipeline-times} presents the execution times for both Phi-2 and Falcon-7B systems under various configurations. These measurements were taken at the optimal $k$ values for each system ($k=13$ for Phi-2, $k=3$ for Falcon-7B), representing the best trade-off between accuracy and inference time in our experiments.

\begin{table}[ht]
\centering
\caption{QA Pipeline Inference Time Results}
\label{tab:qa-pipeline-times}
\begin{tabular}{|c|c|c|}
\hline
Model & Context Chunks ($k$) & Inference Time (s) \\
\hline
Phi-2 & 13 & 2.102 \\
Phi-2 & 0 & 0.524 \\
\hline
Falcon-7B & 3 & 5.023 \\
Falcon-7B & 0 & 4.682 \\
\hline
\end{tabular}
\end{table}

The pipeline execution times reveal significant differences between the two systems. The Phi-2-based system demonstrates faster execution times despite using FP32 precision, compared to Falcon-7B's FP16. This disparity is attributed to several factors:

\begin{itemize}
    \item Model size: Falcon-7B, being substantially larger, requires more computational resources for inference.
    \item Token generation limits: 
    \begin{itemize}
        \item 10 tokens for Phi-2 - since we fine-tuned it to generate the option within the first few tokens.
        \item 100 tokens for Falcon-7B - since the free-form response was often more verbose.
    \end{itemize}
    \item Context utilization: The addition of context chunks ($k$) increases execution time for both systems, with Phi-2 showing a more pronounced relative impact.
\end{itemize}

For the Falcon-7B system, we additionally measured the time for the ensemble scoring process, which adds a relatively small overhead of approximately 49 milliseconds. This step is unique to the Falcon-7B pipeline and contributes to its overall execution time. 

Our approach introduces additional parameters beyond the core models. The embedding model for scoring in the Falcon-7B system adds 434.1M parameters, while the LoRA adapter for Phi-2 introduces 26.2M parameters. However, it's worth noting that the LoRA adapter's parameter count is relatively small compared to the base model size, and does not significantly impact the inference time. These additional parameters contribute to the overall system complexity but play crucial roles in enhancing performance and enabling domain-specific adaptations.

\subsection{Error Analysis}

To categorize the errors made by our Phi-2 QA system, we calculated the error rates for each dataset category (standards specifications and standards overview). Here, the error set refers to the set of questions from the public evaluation set that our Phi-2 system answered incorrectly. We expected that a well-represented category with a notable relative increase in its error set distribution, compared to its dataset distribution, would indicate areas where the model struggles. However, as seen in Table \ref{tab:subdomain-frequencies-vertical}, the observed differences are not significant enough to suggest that any particular category presents a consistent challenge for the model.

\begin{table}[ht]
\centering
\caption{Frequencies per Category: Dataset vs Error Set}
\label{tab:subdomain-frequencies-vertical}
\begin{tabular}{|l|c|c|}
\hline
 & Standards Overview & Standards Specifications \\
\hline
Dataset Count & 63 & 303 \\
Error Set Count & 9 & 65 \\
Dataset \% & 17.2\% & 82.8\% \\
Error Set \% & 12.2\% & 87.8\% \\
Relative \% Change  & -29.3\% & 6.1\% \\
\hline
\end{tabular}
\end{table}

Stronger conclusions can be drawn from the number of questions for which at least one relevant answer-bearing passage was retrieved. As shown in Table \ref{tab:binary-recall-frequencies-vertical}, for half of the questions where the Phi-2 QA system failed, no relevant passage was found. This suggests that the main bottleneck in our system is the retrieval pipeline. 
\begin{table}[ht]
\centering
\caption{Frequencies for Binary Recall: Dataset vs Error Set}
\label{tab:binary-recall-frequencies-vertical}
\begin{threeparttable}
\begin{tabular}{|l|c|c|}
\hline
 & 0 & 1 \\
\hline
Dataset Count\tnote{1} & 24 & 98 \\
Error Set Count & 37 & 37 \\
Dataset \% & 19.7\% & 80.3\% \\
Error Set \% & 50.0\% & 50.0\% \\
Relative Change (\%) & 154.2\% & -37.8\% \\
\hline
\end{tabular}
\begin{tablenotes}
    \item[1] Here, dataset refers to the randomly selected subset of questions for which binary recall was human evaluated.
\end{tablenotes}
\end{threeparttable}
\end{table}

\section{Conclusion}
We have presented QA systems that significantly enhance the performance of small language models on question-answering tasks in the telecommunication domain. By employing a combination of approaches, including model-specific optimizations, ColBERT-based retrieval-augmented generation (RAG), and domain-specific lexical enhancements, we have demonstrated substantial improvements in the language models' abilities to handle technical queries. Our systems achieved 81.9\% accuracy for Phi-2 and 57.3\% for Falcon-7B on the private evaluation dataset. These results underscore the importance of specialized retrieval methods for enhancing the knowledge available during inference, as well as model-specific reasoning optimizations for domain-specific tasks.

\section{Future Work}
There are several promising avenues for future research that could further extend and refine our findings:

\begin{itemize}
    \item Conduct a more comprehensive analysis of the retrieval pipeline:
    \begin{itemize}
        \item Extend our preliminary binary precision analysis at $k=13$ to various $k$ values.
        \item Evaluate additional metrics such as mean reciprocal rank, precision, and recall at different $k$ values to gain deeper insights into retrieval performance.
    \end{itemize}
    
    \item Explore dynamic context selection by resizing context based on query relevance, using a learnable scoring mechanism, potentially optimized via Reinforcement Learning from Human Feedback (RLHF).
    
    \item Improve coherence in retrieved chunks by exploring context stitching techniques to maintain logical order and enhance the quality of the provided context.
    
    \item Incorporate multi-modal input handling by developing document parsers capable of processing diagrams, thereby expanding the range of information our system can utilize.

\end{itemize}



\bibliographystyle{IEEEtran}
\bibliography{references}
\end{document}